\documentclass[letterpaper, 10 pt, journal, twoside]{ieeetran}
\usepackage{amsmath,amsfonts}
\usepackage{algorithmic}
\usepackage{algorithm}
\usepackage{array}
\usepackage[caption=false,font=normalsize,labelfont=sf,textfont=sf]{subfig}
\usepackage{balance}
\usepackage{textcomp}
\usepackage{stfloats}
\usepackage{url}
\usepackage{verbatim}
\usepackage{graphicx}
\usepackage{booktabs}
\usepackage{cite}
\usepackage{xcolor}
\def\eg{\emph{e.g., }} 

\def\ie{\emph{i.e., }}

\def\m{\emph{M }}
\def\M{\emph{M}}

\begin{document}

\title{Introducing \textit{M}: A Modular, Modifiable Social Robot}

\markboth{Antony \MakeLowercase{\textit{et al.}}: M Robot}
{Antony \MakeLowercase{\textit{et al.}}: M Robot} 

\author{Victor Nikhil Antony$^{1*}$, Zhili Gong$^{2*}$, Yoonjae Kim$^{1}$ and Chien-Ming Huang$^{1}$%

\thanks{$^{1}$Department of Computer Science, Johns Hopkins University, USA}
\thanks{$^{2}$Department of Mechanical Engineering, Rice University, USA}
\thanks{$^{*}$Equal Contribution}
}



\maketitle

\begin{abstract}
We present \M, an open-source, low-cost tabletop social robot platform aimed at lowering barriers to social robotics research by making robots easier to reproduce, modify, and deploy in real-world settings. \m combines a modular mechanical design, multimodal sensing, and expressive yet mechanically simple actuation architecture with a ROS2-native software package that cleanly separates perception, expression control, and data management. The platform includes a simulation environment with interface equivalence to hardware to support easy sim-to-real transfer of interaction behaviors. We demonstrate extensibility through additional sensing/actuation modules and provide example interaction templates for storytelling and two-way conversational coaching. Lastly, we report real-world use of M in interaction design and week-long in-home deployments, showing how \m can serve as a practical platform for longitudinal, reproducible social robotics research.
\end{abstract}

\begin{IEEEkeywords}
Social Robotics, Socially Assistive Robotics, Open Source Robot, Mechanical Design, Research Platform.
\end{IEEEkeywords}

\section{Introduction}
Social robots aim to influence human behavior through real-time, socially engaging interactions rather than by performing physical tasks. Using channels such as emotive speech and expressive motion, social robots can coach skills, motivate adherence, and sustain user engagement over repeated interactions. Prior work has demonstrated the potential of social robots in diverse domains, including education \cite{belpaeme2018social}, interventions for children with autism spectrum disorder \cite{scassellati2018improving}, physical activity promotion for older adults \cite{antony2024designing}, and support for healthy routines such as sleep hygiene \cite{antony2025social}. 

Despite these promising applications, enabling truly socially intelligent behavior in robots remains a long-standing grand challenge in robotics. Humans effortlessly perceive, interpret, and respond to rich social signals such as facial expressions, body language, and vocal intonation, whereas robots continue to struggle with robust social perception, reasoning, and expressive behavior \cite{yang2018grand}. Social robots can serve as a critical testbed for addressing this challenge, offering insight into how robots operate fluently in human environments.

In practice, however, social robotics research has progressed slowly, hindered by both fundamental challenges in social intelligence and practical barriers in platform infrastructure. For social robots to be useful in real-world settings, they must produce compelling, believable social behaviors; yet, traditional approaches rely on hand-crafted animations and pre-scripted interactions that feel static and fail outside narrow contexts, limiting effectiveness in longitudinal deployments where meaningful effects are observed. These challenges are compounded by platform friction: many existing social robots are expensive, closed-source, and difficult to reproduce or extend, requiring substantial upfront engineering effort before researchers can investigate interaction strategies. Limited modularity hampers efforts to swap sensors or test alternative expressive capabilities, while commercial platforms constrain long-term maintainability, a fraught dependency made concrete by the repeated discontinuation of widely used research platforms. These failures are not incidental since HRI remains tied to economic models premised on continual growth and market expansion, leaving the field's platform choices contingent on decisions made for reasons unrelated to research \cite{thunberg2026post}. As a result, social robotics research often progresses through isolated, non-replicable platforms, making it difficult to benchmark progress, share advances across research groups, and build on prior work, further hindering the field's ability to conduct the longitudinal, in-the-wild studies where generalizable insights into long-term HRI can be gained.




\begin{figure}[t!]
\centering
\includegraphics[width=\columnwidth]{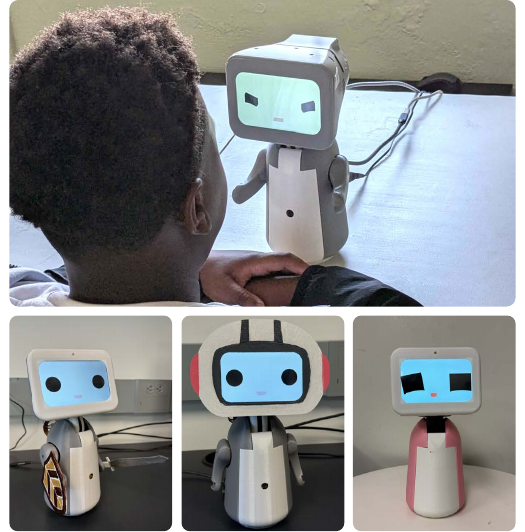}
\caption{\m is an open-source, modular, tabletop social robot platform designed for longitudinal, in-the-wild research. The platform features customizable embodiment (\eg accessories, face plate), expressive behaviors (\eg facial expressions, body gestures), multimodal sensing (\eg touch, radar, camera), and ROS2 software supporting reproducibility, extensibility and deployments.} 
\label{fig:teaser}
\end{figure}


To support scalable research in social robotics, we introduce \textit{M}, an open-source research platform\footnote{project site: \url{https://intuitivecomputing.github.io/M-robot/}} designed to be modular, modifiable, and low-cost, while remaining suitable for real-world deployments (See Fig. \ref{fig:teaser}). \textit{M} couples a modular mechanical design and multimodal sensing with expressive capabilities and a flexible software architecture that supports rapid development, easy reproduction, and simulation-to-hardware transfer. We present the design and implementation of \textit{M}, and detail its feasibility for real-world deployments, positioning \textit{M} as a practical robot platform for longitudinal, in-the-wild social robotics research. 


\section{Related Works}

\subsection{Long-Term, In-the-Wild Social Robotics}
Key effects of social robots (\eg sustained engagement, rapport formation, behavior change) emerge largely through long-term interaction, as user expectations and engagement dynamics evolve over weeks or months \cite{matheus2025long}. Field studies span education in schools \cite{belpaeme2018social}, home-based autism interventions \cite{scassellati2018improving}, storytelling and early language learning \cite{antony2026ella, kory2013robotic}, care environments \cite{carros2020exploring}, and wellbeing coaching \cite{jeong2023deploying, jeong2022designing}, yet each domain has come to depend on a small number of platforms that are either commercially fragile (Nao, Jibo) or confined to their originating lab (Tega) \cite{westlund2016tega}, limiting replication and forcing each study to re-solve the same engineering problems.

These deployments also expose system-level challenges. Long-term studies report substantial effort maintaining hardware reliability, adapting sensing and interaction capabilities as protocols and study goals evolve, and managing logistical and privacy constraints in real settings \cite{matheus2025long, carros2020exploring}. Reconfiguration for each distinct domain typically requires mechanical or architectural rework; remote maintenance is rarely supported, so faults demand physical visits; and logging, monitoring, and privacy handling need to be assembled per requirements for each study. Reconfigurability without rework, sensing breadth, expressivity at sustainable mechanical cost, reproducibility across sites, and unsupervised, autonomous operation motivate M's design principles.

\subsection{Open-Source Robot Platforms}

Open-source social robots have expanded access to embodied HRI research, yet existing platforms make different trade-offs across cost, sensing, expressivity, on-board computation, simulation support, and field deployability rather than offering an integrated research infrastructure (Table I).

Blossom \cite{suguitan2019blossom}, Ono \cite{vandevelde2013systems}, and FLEXI \cite{alves2022flexi} offer affordability ($250–$2.5K) and customizable embodiment. Blossom's fabric-based design supports rapid appearance iteration, while Ono and FLEXI offer mechanical expressivity through multi-DoF actuation, and FLEXI provides on-board compute via an integrated tablet. However, these platforms offer limited multimodal sensing (camera and audio only, or none at all), lack simulation environment for programming, and are not natively architected for autonomous, long-term field deployment with logging, monitoring, or privacy-aware operation.

Humanoid platforms such as Poppy-H \cite{lapeyre2014poppy} and Reachy-Humanoid \cite{mick2019reachy} offer richer sensing, higher degrees of freedom, and ROS compatibility, but are comparatively expensive and mechanically complex, increasing maintenance burden for longitudinal studies. Quori \cite{specian2021quori} is the closest precedent to our goals: an affordable, community-informed platform explicitly designed for non-contact HRI research and distributed to research groups to build a shared evidence base. Its scale, however, depends on dedicated external funding for fabrication and distribution, and it is unclear whether its mobile-base humanoid form can sustain in-home deployment.

A parallel tradition of self-made platforms is built and maintained entirely in-house. Kaspar \cite{wood2021developing} illustrates both the value and the cost of this route: a decade of iterative, user-driven refinement produced a platform well matched to its intervention domain, but one whose design knowledge remains concentrated in a single group and is not straightforward to reproduce elsewhere. Recent compact platforms such as Reachy Mini instead prioritize accessibility and AI integration, lowering cost barriers while maintaining head-based expression, but provide limited sensor extensibility and only partial simulation–hardware parity, well suited for interactive demos and AI experimentation, but not structured as longitudinal research infrastructure. Underlying these efforts is a broader question about the economics of research infrastructure: HRI remains tied to models premised on continual growth and market expansion \cite{thunberg2026post}, and platforms (\eg Furhat, QTrobot, Misty II, Pepper) that depend on commercial viability inherit that fragility regardless of how open their interfaces are.

In contrast, M is designed as a modular and reproducible platform for social robotics research. Rather than maximizing degrees of freedom, M balances expressive minimalism with integrated sensing, ROS-native architecture, and strict simulation–hardware interface equivalence, at a cost profile and reproduction path that do not depend on centralized fabrication.




\begin{table}[t]
\centering
\caption{Comparison of Social and Expressive Robot Platforms}
\label{tab:robot_comparison_m}
\scriptsize
\begin{tabular}{@{}l@{\hspace{3pt}}r@{\hspace{2pt}}c@{\hspace{0pt}}c@{\hspace{0pt}}c@{\hspace{0pt}}c@{\hspace{0pt}}c@{}}
\toprule
\textbf{Robot} & 
\textbf{Cost} & 
\rotatebox{60}{\textbf{Multi-Sensor}} & 
\rotatebox{60}{\textbf{Expressivity}} & 
\rotatebox{60}{\parbox{1.5cm}{\centering\textbf{On-Board}\\\textbf{Compute}}} & 
\rotatebox{60}{\textbf{Sim2Real}} & 
\rotatebox{60}{\parbox{1.5cm}{\centering\textbf{Field}\\\textbf{Readiness}$^*$}} \\ 
\midrule
Blossom \cite{suguitan2019blossom} & \$250 & $\circ$ & 4 DoF & - & - & $\circ$ \\
Poppy-H \cite{lapeyre2014poppy} & \$8-10K & $\bullet$ & 25 DoF, LCD & - & $\bullet$ & - \\
Ono \cite{vandevelde2013systems} & \$350-575 & - & 13 DoF & - & - & - \\
FLEXI \cite{alves2022flexi} & \$2.5K & $\circ$ & 4 DoF, tablet & $\bullet$ & - & $\circ$ \\
Reachy-Mini & \$299-449 & $\bullet$ & 6 DoF & - & $\circ$ & $\circ$ \\
\midrule
\textit{M (ours)} & \textbf{\$325} & $\bullet$ & \textbf{5 DoF, LCD} & $\bullet$ & $\bullet$ & $\bullet$ \\
\bottomrule
\end{tabular}

\smallskip
\par
\footnotesize
\textbf{Legend:} 
$\bullet$ = Full, 
$\circ$ = Partial, 
-- = Not provided. 
\textbf{Multi-Sensor:} $\bullet$ indicates multiple sensing modalities; 
$\circ$ indicates a single sensing modality. 
\textbf{Sim2Real:} $\bullet$ indicates a supported simulation environment with equivalent software/control interfaces to the physical robot; 
$\circ$ indicates simulation support without full interface equivalence. 
\textbf{Field Readiness:} assessed from four deployment-oriented capabilities: integrated data logging, physical privacy controls, containerized runtime, and remote monitoring; 
$\bullet$ indicates all four capabilities, 
$\circ$ indicates a subset, and 
-- indicates none.
\end{table}

\begin{figure}[b!]
\centering
\includegraphics[width=\columnwidth]{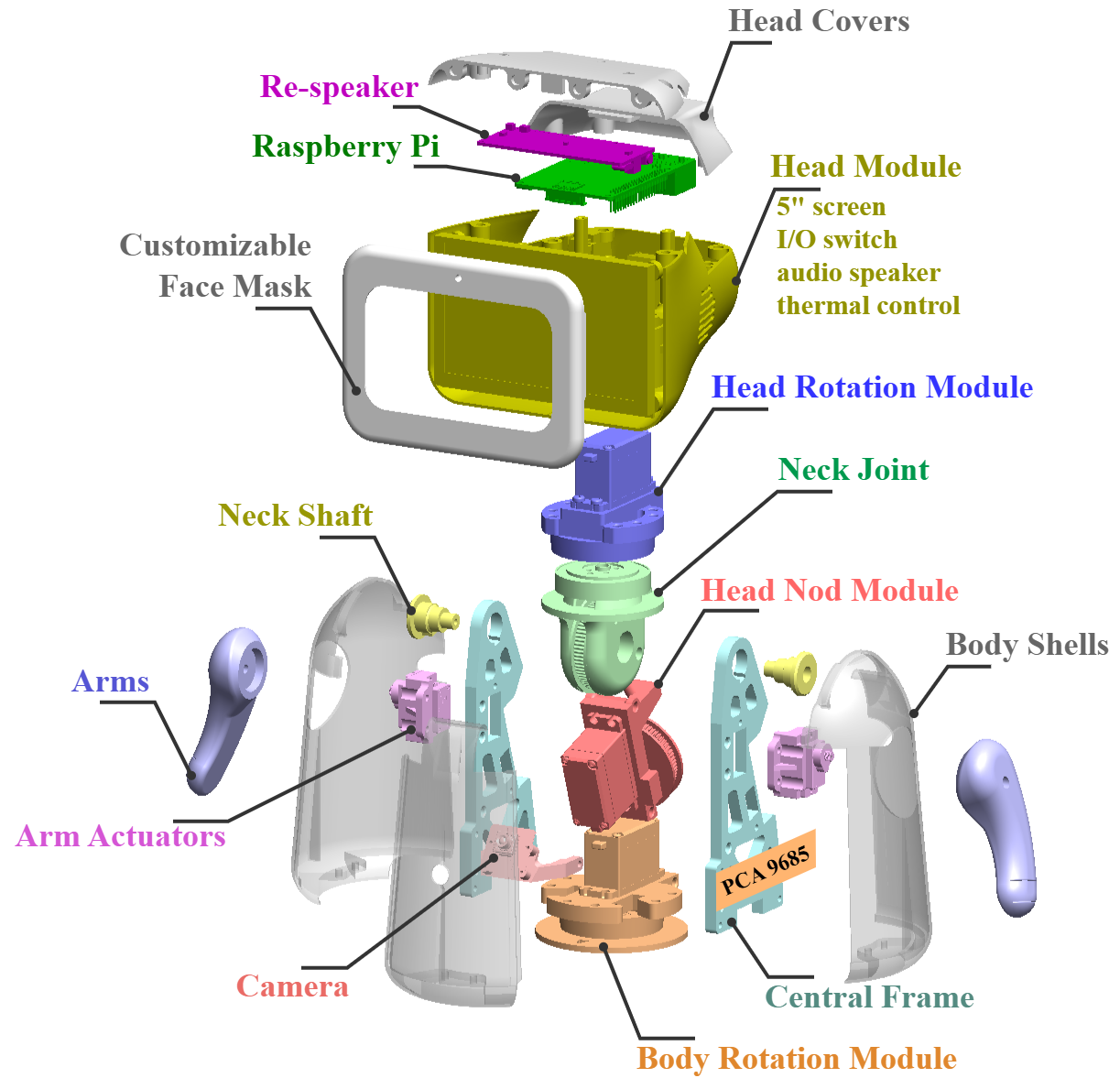}
\caption{Exploded view of \M's mechanical design, illustrating the modular architecture. Key components include the head module (containing compute, audio, and display), head rotation and nodding modules, customizable face masks, articulated arms with magnetic attachments, camera, and body rotation module with weighted base for stability.}
\label{fig:mechanical-design}
\end{figure}

\section{System Overview}
\subsection{Design Rationale}

\begin{figure*}[h]
\centering
\includegraphics[width=\textwidth]{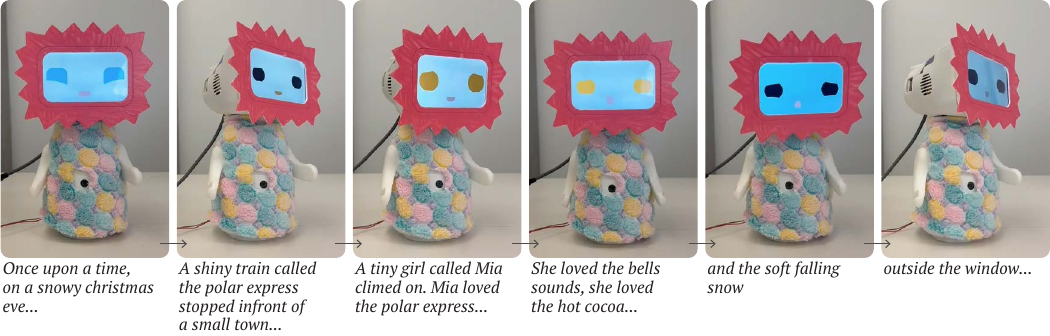}
\caption{Illustration of \m narrating a children's story with synchronized multimodal expressive behaviors produced leveraging generative AI pipelines.}
\label{fig:storytelling}
\end{figure*}


\textbf{Modularity.} To address the limited extensibility of existing platforms (\S II), \m uses modular sensor, shell, and actuation interfaces that allow components to be changed without redesigning the full system. This aims to reduce engineering overhead for iteration as research questions evolve, supporting exploration of embodiment, sensing and interaction configurations. This extensibility is illustrated by the radar and touch sensing modules in \S IV-A and by rapid interaction refinement in the real-world studies in \S V-A.


\textbf{Multimodal Interaction.} Existing open platforms often provide limited sensing or integrate perception and expression only loosely, constraining investigation of coordinated social behavior. M integrates multiple sensing and expressive channels through common software interfaces. \S IV-B and \S IV-C demonstrate this coordination through synchronized speech, expressions, and gestures in storytelling and conversational coaching, while \S IV-A illustrates more sensing modalities.


\textbf{Expressive Embodiment.} Richly articulated robots can provide expressive richness but introduce cost, mechanical complexity, and maintenance burden that hinder longitudinal field deployment. M instead uses five expressive degrees of freedom (head orientation, body orientation, arm gestures) together with facial animation to support rich expressivity while retaining mechanical simplicity. These implemented behaviors are exercised in the storytelling and conversational coaching demonstrations (\S IV-B–C; Figs. \ref{fig:storytelling} and \ref{fig:positive-psychology}).


\textbf{Reproducibility.} Platform cost, complexity, and inconsistent hardware/software environments make social robotics studies difficult to reproduce across researchers and sites. M targets these barriers through cost-accessible components, documented mechanical designs, defined ROS 2 interfaces, containerized software, and simulation–hardware interface equivalence. The educational deployment in \S V-B illustrates this workflow, where the same containerized simulation environment is used before validation on physical robots.


\textbf{Deployability.} Long-term studies report that engineering effort shifts toward maintenance, privacy management, and logistics once robots leave the lab. M aims to address these needs through tool-free mechanical access, optional camera covers and radar sensing for privacy-centric deployments, containerized runtimes, session-level logging, and remote monitoring. These mechanisms were exercised in week-long home deployments reported in \S V-A, which used autonomous operation, interaction logging, and remote usage monitoring.


\subsection{Mechanical Design}

The mechanical design of \m centers on modularity, ease of modification, and support for expressive motion while maintaining mechanical simplicity and robustness \footnote{see BoM: https://intuitivecomputing.github.io/M-robot/bill-of-materials}. The robot is composed of two primary assemblies, the head and the body, each designed to house specific actuation, sensing, and computation components while enabling straightforward access, clean wire management, and physical customization. \m uses hobby-grade microservos for arm actuation and higher-torque 20 kg·cm servos for all other actuated joints.

\textbf{Head Assembly.} The head assembly houses the primary computational, audio, and thermal management components, as well as the yaw actuator used for expressive head motion (see Fig. \ref{fig:mechanical-design}). Specifically, the head contains the head yaw motor, an onboard single-board computer (Raspberry Pi), a ReSpeaker audio module providing both microphone and speaker functionality, an internal cooling fan for heat dissipation, and an external I/O power switch. These components are mounted within an internal holder that secures both the motor and compute hardware, providing structural stability while maintaining accessibility for maintenance and replacement.

The external head shell is divided into four primary components to support modularity and customization. The left and right base shells enclose the internal cavity and provide structural support. A removable face plate encloses the display and connects the base shells, allowing rapid replacement for aesthetic or functional customization without disassembling the full head assembly. The ReSpeaker module is mounted to the front top cap, which is mechanically fastened to the base shell to ensure stable acoustic positioning; the rear-mounted speaker also contributes to anchoring the left and right base shells. A rear top cap is magnetically attached, enabling tool-free access to internal components for debugging, maintenance, or hardware modification. This combination of mechanically fastened and magnetically attached elements balances rigidity with ease of access.

\textbf{Body Assembly.} The body assembly consists of a central structural frame that supports the primary actuators and connects the head and arm subsystems. The frame houses the head pitch motor, which is mechanically coupled to the head module via a pulley-based transmission. This configuration enables expressive pitch motion while isolating the motor from the head enclosure, reducing inertial load on the head and simplifying both mechanical routing and cable management.
Two articulated arms are mounted on either side of the central frame and actuated using micro-servo motors. Each arm incorporates embedded magnets to support rapid attachment of accessories or end-effectors, enabling physical customization without mechanical rework. A PCA9685 motor control board is mounted on the the central frame and interfaces with all servo motors in the system. This centralized motor control architecture simplifies wiring by routing most of servo connections within the body, rather than directly to the compute module in the head. As a result, only two cables exit the body toward the head assembly: a 27W USB-C power cable supplying the onboard compute and a 9V DC power cable connecting to the PCA board. This design aims to reduce cable clutter, improve reliability during motion, and facilitate maintenance and robustness during deployment.

The body shell is divided into three segments: left, right, and center plates. The left and right shell segments include apertures that allow the arm linkages to connect directly to their respective motors while maintaining enclosure continuity. A camera mount can be attached to the central frame, positioning the camera to peer through an aperture in the central body shell for forward-facing visual sensing which can be optionally covered with a magnetically attached physical-privacy cover.  The central shell plate provides access points for additional sensors, supporting extensibility without requiring redesign of the primary structural components. All shell segments are mechanically fastened to the frame at the base, ensuring structural stability while allowing disassembly when needed.

The base yaw motor is coupled to a dedicated base yaw module that serves as a rotational anchor point. This module incorporates weighted elements and anti-slip pads to improve stability during operation and expressive motion. Together, these design choices support expressive articulation while maintaining a compact, stable, and maintainable mechanical structure suitable for extended deployment (see Fig. \ref{fig:storytelling}).


\subsection{Software Architecture}
\M's software architecture is designed to support extensible, multimodal socially intelligent interaction while remaining modular, reproducible, and deployable across research contexts. Rather than coupling interaction logic to specific algorithms or hardware implementations, the system is organized around a set of core functional components with clear interfaces. These interfaces are formalized through ROS2 message, service, and action definitions that specify data formats and control semantics, separating \textit{what} each capability provides from \textit{how} it is implemented and enabling components to be modified or replaced without restructuring the overall system.

At a high level, the core architecture comprises three interacting components: (1) multimodal perception, (2) expression control, (3) data management. Components communicate using standard ROS 2 patterns (\ie topics for streaming data, services for synchronous requests, and actions for long-running or interruptible operations) supporting loose coupling and incremental system evolution.

\textbf{Multimodal Perception.} \M's perception packages provide access to audio, video, and other sensors (\eg mmwave radar, touch) for the rest of the program. Sensor streams are exposed as continuous topics and processed by independent perception modules responsible for tasks such as speech activity detection, speech recognition, visual sensing, or environmental monitoring. Decomposing perception into parallel, replaceable stages enables experimentation with alternative algorithms and sensing modalities without requiring behavior logic changes.

\textbf{Expression Control.} M represents expressive behaviors as timed sequences of actions. Rather than directly specifying low-level commands (\ie individual motor positions), interaction programs call higher-level behaviors that define what the robot should do and when. For e.g., a behavior can specify a head turn, arm gesture, or facial expression along with its timing. The controller then handles the corresponding motor and display commands, including smooth transitions and interruption when another behavior needs to take over. This separation allows researchers to coordinate expressive behaviors without managing low-level actuation directly.

\textbf{Data Management}. \m provides infrastructure for session-level tracking and multimodal data logging essential for longitudinal social robot deployment. The \textit{m-logging} package supports synchronized recording of sensor streams, system events, and interaction state, enabling researchers to monitor deployment health, validate interaction execution, and reconstruct user experiences post-deployment. Data streams are structured with timestamps and metadata to support longitudinal analysis and identification of technical issues during extended field studies. Remote monitoring capabilities allow non-intrusive verification of system operation during deployments.


\subsection{Simulation}
\m includes a simulation environment designed to support rapid iteration, safe development, and reproducible transfer of interaction behaviors from simulation to physical deployment.

A key design principle of the simulation environment is interface equivalence with the physical system. The simulated robot exposes the same multimodal perception topics, expression control interfaces, and data management mechanisms as the physical robot. Hardware-specific components (such as motors, displays, sensors) are replaced by simulated counterparts that implement identical ROS 2 message, service, and action interfaces. This consistency allows perception pipelines, behavior coordination logic, and expressive behaviors to be developed and tested in simulation without modification prior to deployment on the physical robot.

\begin{figure}[h]
  \includegraphics[width=\columnwidth]{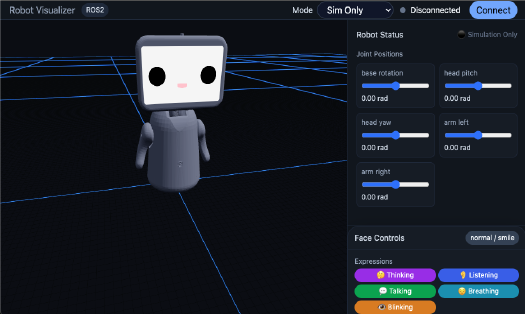}
  \caption{\m can be programmed in a simulation environment with complete sim-to-real parity. The GUI includes a 3D viewport with orbit controls, real-time joint state displays, and interactive manipulation controls.}
  \label{fig:simulation}
\end{figure}

The simulation supports development of complete interaction loops, not only expressive motion. In addition to simulated actuation and display outputs, the environment provides access to audio and video streams from the host machine, enabling testing of speech-based interaction, multimodal perception, and closed-loop interaction logic under conditions that closely mirror real-world operation. This allows researchers to prototype, debug, and evaluate interaction behaviors that integrate perception, reasoning, and expression without the hardware.

To support observability during development and deployment, the simulation provides a real-time, web-based visualization interface for inspecting robot state and interaction execution (see Fig. \ref{fig:simulation}). The interface renders a URDF model of \m in an interactive viewport with orbit-style camera control and a ground reference grid, allowing experimenters to inspect kinematic behavior from consistent viewpoints. In simulation-only mode, \M's expressive degrees of freedom (base rotation, head pitch and yaw, and bilateral arms) can be directly manipulated through bounded controls. In mirroring mode, the same visualization becomes read-only and instead reflects live joint states streamed from the physical robot, providing a digital twin view for remote monitoring and debugging

By minimizing divergence between simulation and deployment, \m supports a workflow in which interaction behaviors can be iteratively developed, inspected, and refined in simulation and then transferred to the physical robot. This approach aims to reduce engineering overhead, improve safety during development, and support collaboration and reproducible experimentation across environments and research groups.

\section{Demonstrating Usage}
\subsection{Extensions} To demonstrate \M's modularity and extensibility,  we integrated additional sensing and actuation modalities and illustrate the potential use cases below.


\textit{Non-Instrusive Human-Presence Detection.} \m supports a Frequency Modulated Continuous Wave (FMCW) radar sensor mounted on its front plate, enabling detection of human presence and coarse motion without capturing visual or audio data. This privacy-preserving modality allows \m to remain aware of its social context without relying on cameras or microphones. \textit{Imagine} a daily mental health check-in scenario: upon detecting a user’s presence, \m may respond with subtle, playful behaviors (\eg gently reorienting its body, animating its display) to proactively invite engagement and reinforce its social presence. These low-stakes initiation cues are particularly valuable for interventions that depend on consistent participation, allowing \m to remain socially available while respecting the rhythms of everyday life.

\textit{Capacitive Touch Sensing.} Capacitive touch sensors embedded beneath \M’s exterior shell enable the robot to sense direct physical contact through its surface. This capability supports interactions in which touch functions as an intentional yet lightweight signal such as acknowledging attention, requesting continuation, or expressing reassurance. For example, a brief tap on the robot’s shell may signal readiness to engage, while sustained contact may be requested as part of the interaction. By integrating touch sensing beneath the shell, \m enables designing embodied exchanges without introducing additional visible instrumentation or altering its external form.


\textit{Vibro-Tactile Actuation.} \M's shell can have linear resonant actuator (LRA) vibration motors that provide localized vibro-tactile feedback through its shell. This modality supports interactions grounded in rhythm and bodily sensation rather than visual or auditory cues. Imagine a guided breathing exercise in which a user rests their hands on the robot, while subtle, rhythmic vibrations cue inhalation and exhalation, reinforcing calm, paced breathing through touch. This form of feedback introduces an unobtrusive expressive channel that is well-suited to low-arousal or therapeutic interactions, complementing \M’s existing visual and auditory capabilities.

\textbf{On Privacy.} M is not privacy-preserving in every configuration; instead, its modularity allows each deployment to select modalities appropriate to its privacy requirements. In privacy-sensitive settings, radar could provide coarse presence and motion sensing without capturing audio or video, while physical camera covers offer an explicit visual-sensing control.

\subsection{Example Interaction: One-Way Storytelling}

To illustrate how rich, expressive multimodal behaviors can be delivered with \M, we include an example storytelling interaction in which \m narrates a complete story with synchronized expressive behaviors (\ie body gestures, emotive speech and facial animations). For instance, when narrating a winter tale, \m displays wide-eyed wonder during magical moments, tilts its head thoughtfully during reflective passages, and uses its arms to emphasize key story beats\footnote{link to demo video: \url{https://youtube.com/shorts/iwfRNtz6DAI}} (see Fig. \ref{fig:storytelling}).

This storytelling interaction is implemented as an example ROS 2 package that demonstrates how expressive behaviors can be authored, scheduled, and executed on \M. The story and its behaviors are generated with an LLM-based pipeline that decomposes the narrative into sequential chunks, each paired with corresponding expressive cues\cite{antony2025xpress}. These cues specify gesture animations and timing information relative to spoken audio, allowing expressions to be synchronized with narration without embedding low-level control logic into the interaction code. At runtime, a dedicated delivery node manages story progression using a state machine, coordinating audio playback via \M's ROS 2 action interfaces and scheduling expressive behaviors through \M's animation execution framework.

Although the storytelling interaction is intentionally one-way, the same structure can be extended to support conditional branching, user-responsive behaviors, or multimodal interaction, making it a practical reference for researchers developing more complex expressive engagements with \M.

\begin{figure*}[t]
\centering
\includegraphics[width=\textwidth]{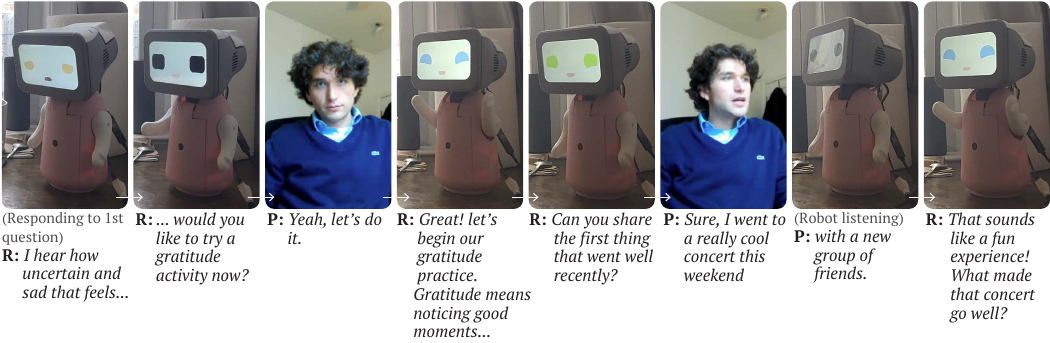}
\caption{Illustration of \m delivering a two-way conversational positive psychology coaching session. \m guides a structured gratitude practice through open-ended questions, empathetic listening with embodied acknowledgment, and context-aware follow-up dialogue maintained across multiple conversational turns.}
\label{fig:positive-psychology}
\end{figure*}

\subsection{Example Interaction: Two-Way Conversation}

To illustrate two-way, socially grounded interaction, \m includes an example conversational positive psychology coaching interaction in which the robot guides a five-day structured therapeutic activity \cite{jeong2020robotic}. In each daily session, \m engages in a targeted practice session (see Fig. \ref{fig:positive-psychology}): the robot asks open-ended questions about positive experiences, listens to and acknowledges their responses with empathetic body language and facial expressions, and asks follow-up questions that build on what the user has shared. Throughout the conversation, \m maintains context-appropriate dialogue over multiple turns, adapting its responses based on what the user says while tracking the session's therapeutic goals.

User speech is processed into conversational turns that are tracked by a turn manager and session state tracker, which together maintain context across the interaction, including dialogue history, conversational phase, and task progress. A response generation module uses this state to produce context-appropriate utterances and select facial expressions and body gestures using a large language model. Generated responses are represented as structured conversational acts that pair spoken content with facial expressions and body gestures.

A dedicated delivery component then executes these responses by coordinating speech output and embodied expression through \m’s animation and action interfaces. By separating conversational state management, response generation, and expressive execution, this template enables iterative refinement of conversational behavior without entangling dialogue logic with low-level control. While the included example focuses on a guided, therapeutic-style coaching session, the same interaction implementation pattern can generalize  to other forms of dialogue, such as check-ins, reflective conversations, or mixed-initiative interactions, providing a reusable scaffold for building interactive social robotic interventions on \M.



\section{\m in the Real-World}

To illustrate evidence of \M's ability to facilitate social robotics research, we report on real-world use of the platform across research and educational contexts. These demonstrate that \M's capabilities (\ie modular hardware, ROS-native software stack, containerized runtime environment) support longitudinal, unsupervised in-home interactions and rapid iterations.



\begin{figure*}[h]
  \includegraphics[width=\textwidth]{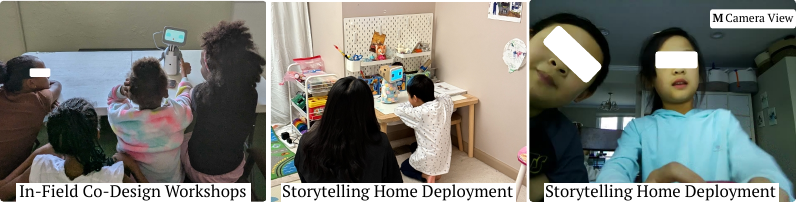}
  \caption{\m in real-world deployment: participatory design workshops (left) and week-long autonomous home deployments (center, right).}
  \label{fig:field_use_cases}
\end{figure*}

\subsection{\m for Child-Robot Interaction}
\m has been extensively used within our research group as a platform for child–robot interaction studies, including both field-based design and longitudinal home deployments (see Fig. \ref{fig:field_use_cases}). In design sessions conducted in community centers and family homes, \m supported iterative prototyping with children and caregivers and rapid refinement of interaction behaviors without mechanical redesign. To explore longitudinal interaction, we developed a generative AI–powered storytelling robot, \textit{ELLA}, designed to support early language development at home with \m as the platform. In week-long deployments of \textit{ELLA} in 10 homes, children engaged with the robot through interactive storytelling sessions incorporating parent-selected vocabulary targets and scaffolded dialogue \cite{antony2026ella}. The system operated autonomously in-home, generating personalized story content and logging interaction transcripts and timestamps. Across deployments, \m maintained stable operation (only one incident of motor failure was observed and the system as a whole continued functioning as a result of its modular architecture) and collected 14 hours of interaction data alongside remote usage monitoring, demonstrating its suitability for longitudinal in-field child–robot interaction research.

\subsection{\m as an Education Platform}
Beyond research, \m is currently used as the primary robot platform in an undergraduate ``\textit{Introduction to Human-Robot Interaction}'' course at a research-intensive university. Students prototype their own robot interactions within a Docker container hosting the simulation that mirrors the physical M. A shared laboratory space equipped with four \m platforms supports students to validate their simulated interactions on the physical robot. Since M has simulation-to-hardware parity, students are able to focus on interaction design, perception pipelines, and experimental methodology rather than low-level hardware control debugging. The containerized workflow also supported reproducibility and collaboration across student teams. In this context, \m functions as practical pedagogical infrastructure for hands-on HRI training.

\section{Discussion}
\m aims to push the frontiers of socially intelligent robotics by lowering the practical barriers in this space. As a modular, low-cost, and customizable platform, \m can enable researchers to move beyond short-term, lab-bound studies toward longitudinal, in-the-wild investigations that are critical for deepening our understanding of social robots. Our experiences developing and deploying \m have surfaced key opportunities and open challenges for the platform and for social robotics research more broadly. Below, we discuss interconnected directions for enabling truly long-horizon deployment, and fostering meaningful community adoption.

\textbf{Designing for No-Contact, Long-Horizon Deployment.}
Although \m supports in-home and field-based HRI, scaling beyond the week-long deployments reported here to multi-year horizons introduces challenges that remain uncharacterized. Across ten home deployments, one motor failure occurred; the rest of the system continued operating, but longer-term uptime, software failures, recovery behavior, and maintenance burden remain open questions. Long-term studies require systems that can be updated and maintained with minimal physical intervention, particularly for non-technical users or special populations \cite{jeong2023robotic}. This requires infrastructure beyond containerization, including remote updates, fault detection and recovery, version control, and backward compatibility. Prolonged deployments also require attention to operational safety, including actuator limits, safe failure behavior, and clear user protocol when faults occur. Human-facing aspects of deployment are equally important. Unpacking, onboarding, and off-boarding should help users understand the robot's capabilities, limitations, and data practices \cite{lee2022unboxing}. \m provides optional camera covers, radar-based presence sensing, interaction logging, and remote monitoring, but longer deployments require clearer policies for consent, data retention and access, storage scalability, and data minimization. Material longevity is another concern: repair, part substitution, and component reuse over multi-year horizons can conflict with replacement-oriented hardware cycles \cite{thunberg2026post}. \M's printed structure and commodity parts support local repair, but sustaining reliable and safe operation over years remains an open challenge.

\textbf{Supporting Meaningful Community Integration.}
Despite numerous open-source robot platforms, few have achieved sustained adoption in HRI. While \m emphasizes reproducibility, modularity, and deployability, technical design alone is insufficient for shared research infrastructure. Our field deployments suggest that ROS2 interfaces, containerized environments, and replaceable components can support reuse, but broader reproducibility will also depend on updated documentation, ease of setup, troubleshooting guidance, versioned design resources, and maintenance expectations. Manufacturing remains a barrier: \m relies heavily on 3D-printed parts, and access to suitable printers, materials, and fabrication expertise cannot be assumed across research groups. Recent work similarly emphasizes reusable HRI software. ROS2-based frameworks reduce dependence on platform-specific APIs \cite{bono2024open}, while emerging extensions standardize integration of additional sensing modalities \cite{scheibl2026sense4hri}. Cloud- and edge-based architectures can offload computationally intensive services, but introduce concerns around connectivity, privacy, and system dependence \cite{eze2023internet}. Embodied conversational systems likewise separate perception, dialogue, memory, and behavior generation into reusable components \cite{kim2024understanding}. For \m, these trends reinforce that community adoption will depend on reusable interfaces and mechanisms for sharing components, interaction code, data, and experimental artifacts. Documentation, transparent adoption tradeoffs, standardized evaluation, and mechanisms for sharing longitudinal datasets will be important if \m is to develop into a truly shared research infrastructure.

\section{Limitations}
\m has limitations that represent opportunities for community development. Long-term durability beyond week-long deployments remains uncharacterized. Servo motors produce audible noise during motion that may affect noise-sensitive interactions. \M's base capabilities do not include advanced social perception or intelligence modules (\eg emotion recognition, gaze tracking, engagement estimation); We view \m as a collaborative research infrastructure that can evolve through shared extensions and improvements, collectively advancing the field's capacity for longitudinal, reproducible research.

\textbf{Acknowledgments:} \textit{Funding.} National Science Foundation award \#2143704.
\textit{Author CRediT.}  Concept. \& Method. (VNA, ZG, CH); Vis.; Investigation; Review \& Editing (all); Original Draft (VNA); Funding \& Supervision (CH).
\textit{AI Usage.} Table formatting and copyediting with all content verified by authors.



\bibliographystyle{IEEEtran}
\bibliography{references}

@article{yang2018grand,
  title={The grand challenges of science robotics},
  author={Yang, Guang-Zhong and Bellingham, Jim and Dupont, Pierre E and Fischer, Peer and Floridi, Luciano and Full, Robert and Jacobstein, Neil and Kumar, Vijay and McNutt, Marcia and Merrifield, Robert and others},
  journal={Science robotics},
  volume={3},
  number={14},
  pages={eaar7650},
  year={2018},
  publisher={American Association for the Advancement of Science}
}

@article{scassellati2018improving,
  title={Improving social skills in children with ASD using a long-term, in-home social robot},
  author={Scassellati, Brian and Boccanfuso, Laura and Huang, Chien-Ming and Mademtzi, Marilena and Qin, Meiying and Salomons, Nicole and Ventola, Pamela and Shic, Frederick},
  journal={Science Robotics},
  volume={3},
  number={21},
  pages={eaat7544},
  year={2018},
  publisher={American Association for the Advancement of Science}
}

@article{belpaeme2018social,
  title={Social robots for education: A review},
  author={Belpaeme, Tony and Kennedy, James and Ramachandran, Aditi and Scassellati, Brian and Tanaka, Fumihide},
  journal={Science robotics},
  volume={3},
  number={21},
  pages={eaat5954},
  year={2018},
  publisher={American Association for the Advancement of Science}
}

@article{antony2025social,
  title={Social Robots for Sleep Health: A Scoping Review},
  author={Antony, Victor Nikhil and Li, Mengchi and Lin, Shu-Han and Li, Junxin and Huang, Chien-Ming},
  journal={International Journal of Social Robotics},
  volume={17},
  number={4},
  pages={763--777},
  year={2025},
  publisher={Springer}
}

@article{antony2024designing,
  title={Designing social robots that engage older adults in exercise: A case study},
  author={Antony, Victor Nikhil and Huang, Chien-Ming},
  journal={arXiv preprint arXiv:2403.04153},
  year={2024}
}

@article{suguitan2019blossom,
  title={Blossom: A handcrafted open-source robot},
  author={Suguitan, Michael and Hoffman, Guy},
  journal={ACM Transactions on Human-Robot Interaction (THRI)},
  volume={8},
  number={1},
  pages={1--27},
  year={2019},
  publisher={ACM New York, NY, USA}
}

@inproceedings{vandevelde2013systems,
  title={Systems overview of Ono: a DIY reproducible open source social robot},
  author={Vandevelde, Cesar and Saldien, Jelle and Ciocci, Maria-Cristina and Vanderborght, Bram},
  booktitle={Social Robotics: 5th International Conference, ICSR 2013, Bristol, UK, October 27-29, 2013, Proceedings 5},
  pages={311--320},
  year={2013},
  organization={Springer}
}

@inproceedings{alves2022flexi,
  title={Flexi: A robust and flexible social robot embodiment kit},
  author={Alves-Oliveira, Patr{\'\i}cia and Bavier, Matthew and Malandkar, Samrudha and Eldridge, Ryan and Sayigh, Julie and Bj{\"o}rling, Elin A and Cakmak, Maya},
  booktitle={Proceedings of the 2022 ACM Designing Interactive Systems Conference},
  pages={1177--1191},
  year={2022}
}

@inproceedings{lapeyre2014poppy,
  title={Poppy project: open-source fabrication of 3D printed humanoid robot for science, education and art},
  author={Lapeyre, Matthieu and Rouanet, Pierre and Grizou, Jonathan and Nguyen, Steve and Depraetre, Fabien and Le Falher, Alexandre and Oudeyer, Pierre-Yves},
  booktitle={Digital Intelligence 2014},
  pages={6},
  year={2014}
}

@inproceedings{jeong2020robotic,
  title={A robotic positive psychology coach to improve college students’ wellbeing},
  author={Jeong, Sooyeon and Alghowinem, Sharifa and Aymerich-Franch, Laura and Arias, Kika and Lapedriza, Agata and Picard, Rosalind and Park, Hae Won and Breazeal, Cynthia},
  booktitle={2020 29th IEEE international conference on robot and human interactive communication (RO-MAN)},
  pages={187--194},
  year={2020},
  organization={IEEE}
}

@inproceedings{antony2025xpress,
  title={Xpress: A system for dynamic, context-aware robot facial expressions using language models},
  author={Antony, Victor Nikhil and Stiber, Maia and Huang, Chien-Ming},
  booktitle={2025 20th ACM/IEEE International Conference on Human-Robot Interaction (HRI)},
  pages={958--967},
  year={2025},
  organization={IEEE}
}

@inproceedings{jeong2023robotic,
  title={A robotic companion for psychological well-being: A long-term investigation of companionship and therapeutic alliance},
  author={Jeong, Sooyeon and Aymerich-Franch, Laura and Alghowinem, Sharifa and Picard, Rosalind W and Breazeal, Cynthia L and Park, Hae Won},
  booktitle={Proceedings of the 2023 ACM/IEEE international conference on human-robot interaction},
  pages={485--494},
  year={2023}
}

@inproceedings{lee2022unboxing,
  title={The unboxing experience: Exploration and design of initial interactions between children and social robots},
  author={Lee, Christine P and Cagiltay, Bengisu and Mutlu, Bilge},
  booktitle={Proceedings of the 2022 CHI conference on human factors in computing systems},
  pages={1--14},
  year={2022}
}

@article{mick2019reachy,
  title={Reachy, a 3D-printed human-like robotic arm as a testbed for human-robot control strategies},
  author={Mick, S{\'e}bastien and Lapeyre, Mattieu and Rouanet, Pierre and Halgand, Christophe and Benois-Pineau, Jenny and Paclet, Florent and Cattaert, Daniel and Oudeyer, Pierre-Yves and de Rugy, Aymar},
  journal={Frontiers in neurorobotics},
  volume={13},
  pages={65},
  year={2019},
  publisher={Frontiers Media SA}
}

@article{matheus2025long,
  title={Long-term interactions with social robots: Trends, insights, and recommendations},
  author={Matheus, Kayla and Ramnauth, Rebecca and Scassellati, Brian and Salomons, Nicole},
  journal={ACM Transactions on Human-Robot Interaction},
  volume={14},
  number={3},
  pages={1--42},
  year={2025},
  publisher={ACM New York, NY}
}

@inproceedings{carros2020exploring,
  title={Exploring human-robot interaction with the elderly: results from a ten-week case study in a care home},
  author={Carros, Felix and Meurer, Johanna and L{\"o}ffler, Diana and Unbehaun, David and Matthies, Sarah and Koch, Inga and Wieching, Rainer and Randall, Dave and Hassenzahl, Marc and Wulf, Volker},
  booktitle={Proceedings of the 2020 CHI conference on human factors in computing systems},
  pages={1--12},
  year={2020}
}

@inproceedings{antony2026ella,
  title={ELLA: Generative AI-powered social robots for early language development at home},
  author={Antony, Victor Nikhil and Cao, Shiye and Wang, Shuning and Huang, Chien-Ming},
  booktitle={Proceedings of the 25th Annual ACM Interaction Design and Children Conference},
  pages={173--193},
  year={2026}
}

@article{jeong2023deploying,
  title={Deploying a robotic positive psychology coach to improve college students’ psychological well-being},
  author={Jeong, Sooyeon and Aymerich-Franch, Laura and Arias, Kika and Alghowinem, Sharifa and Lapedriza, Agata and Picard, Rosalind and Park, Hae Won and Breazeal, Cynthia},
  journal={User Modeling and User-Adapted Interaction},
  volume={33},
  number={2},
  pages={571--615},
  year={2023},
  publisher={Springer}
}

@phdthesis{jeong2022designing,
  title={Designing and Deploying Robotic Companions to Improve Human Psychological Wellbeing},
  author={Jeong, Sooyeon},
  year={2022},
  school={Massachusetts Institute of Technology}
}

@inproceedings{westlund2016tega,
  title={Tega: a social robot},
  author={Westlund, J Kory and Lee, Jin Joo and Plummer, Luke and Faridi, Fardad and Gray, Jesse and Berlin, Matt and Quintus-Bosz, Harald and Hartmann, Robert and Hess, Mike and Dyer, Stacy and others},
  booktitle={International Conference on Human-Robot Interaction},
  year={2016}
}

@inproceedings{kory2013robotic,
  title={Robotic learning companions for early language development},
  author={Kory, Jacqueline M and Jeong, Sooyeon and Breazeal, Cynthia L},
  booktitle={Proceedings of the 15th ACM on International conference on multimodal interaction},
  pages={71--72},
  year={2013}
}

@inproceedings{bono2024open,
  title={Open Access NAO (OAN): a ROS2-based software framework for HRI applications with the NAO robot},
  author={Bono, Antonio and Brameld, Kenji and D’Alfonso, Luigi and Fedele, Giuseppe},
  booktitle={2024 33rd IEEE International Conference on Robot and Human Interactive Communication (ROMAN)},
  pages={522--527},
  year={2024},
  organization={IEEE}
}

@article{scheibl2026sense4hri,
  title={Sense4HRI: A ROS 2 HRI Framework for Physiological Sensor Integration and Synchronized Logging},
  author={Scheibl, Manuel and Leichert, Julian and G{\"o}rmez, Sinem and Wrede, Britta},
  journal={arXiv preprint arXiv:2603.19914},
  year={2026}
}

@article{eze2023internet,
  title={Internet of Things meets robotics: A survey of cloud-based robots},
  author={Eze, Chrisantus},
  journal={arXiv preprint arXiv:2306.02586},
  year={2023}
}

@inproceedings{kim2024understanding,
  title={Understanding large-language model (llm)-powered human-robot interaction},
  author={Kim, Callie Y and Lee, Christine P and Mutlu, Bilge},
  booktitle={Proceedings of the 2024 ACM/IEEE international conference on human-robot interaction},
  pages={371--380},
  year={2024}
}

@inproceedings{thunberg2026post,
  title={Post-growth Perspectives in HRI},
  author={Thunberg, Sofia and Gamboa, Mafalda and Torre, Ilaria and Penzenstadler, Birgit},
  booktitle={Proceedings of the 21st ACM/IEEE International Conference on Human-Robot Interaction},
  pages={1388--1398},
  year={2026}
}

@article{specian2021quori,
  title={Quori: A community-informed design of a socially interactive humanoid robot},
  author={Specian, Andrew and Mead, Ross and Kim, Simon and Matari{\'c}, Maja and Yim, Mark},
  journal={IEEE Transactions on Robotics},
  volume={38},
  number={3},
  pages={1755--1772},
  year={2021},
  publisher={IEEE}
}

@article{wood2021developing,
  title={Developing kaspar: a humanoid robot for children with autism},
  author={Wood, Luke J and Zaraki, Abolfazl and Robins, Ben and Dautenhahn, Kerstin},
  journal={International Journal of Social Robotics},
  volume={13},
  number={3},
  pages={491--508},
  year={2021},
  publisher={Springer}
}

\end{document}